\documentclass[10pt,twocolumn,letterpaper]{article}

\usepackage[pagenumbers]{cvpr} 

\usepackage{graphicx}
\usepackage{amsmath}
\usepackage{amssymb}
\usepackage{booktabs}
\usepackage{multirow}
\usepackage{makecell}
\usepackage{bbm}
\usepackage{mathtools}
\makeatletter
\@namedef{ver@everyshi.sty}{}
\makeatother
\usepackage{tikz}
\newcommand*\circled[1]{\tikz[baseline=(char.base)]{
            \node[shape=circle,draw,inner sep=0.7pt] (char) {#1};}}

\usepackage[pagebackref,breaklinks,colorlinks]{hyperref}

\usepackage[capitalize]{cleveref}
\crefname{section}{Sec.}{Secs.}
\Crefname{section}{Section}{Sections}
\Crefname{table}{Table}{Tables}
\crefname{table}{Tab.}{Tabs.}

\def\cvprPaperID{*****} 
\def\confName{CVPR}
\def\confYear{2022}

\begin{document}

\title{DAMix: A Density-Aware Mixup Augmentation \\for Single Image Dehazing under Domain Shift}

\author{\textbf{Chia-Ming Chang}\textsuperscript{$\ast$} \quad \textbf{Tsung-Nan Lin}\textsuperscript{$\ast\dag$}\\
\textsuperscript{$\ast$}Graduate Institute of Communication Engineering \\
\textsuperscript{$\dag$}Department of Electrical Engineering \\
National Taiwan University\\
{\tt\small \{r09942090, tsungnan\}@ntu.edu.tw}
}
\maketitle

\begin{abstract}
   Deep learning-based methods have achieved considerable success on single image dehazing in recent years. However, these methods are often subject to performance degradation when domain shifts are confronted. Specifically, haze density gaps exist among the existing datasets, often resulting in poor performance when these methods are tested across datasets. To address this issue, we propose a density-aware mixup augmentation (DAMix). DAMix generates samples in an attempt to minimize the Wasserstein distance with the hazy images in the target domain. These DAMix-ed samples not only mitigate domain gaps but are also proven to comply with the atmospheric scattering model. Thus, DAMix achieves comprehensive improvements on domain adaptation. Furthermore, we show that DAMix is helpful with respect to data efficiency. Specifically, a network trained with half of the source dataset using DAMix can achieve even better adaptivity than that trained with the whole source dataset but without DAMix.
\end{abstract}

\section{Introduction}
Artificial intelligence has acted as the primary driver of progress in several popular fields such as computer vision. However, the degradation of information stemming from haze and fog often results in biases in contrast and color fidelity. In addition, contaminated images impose a burden on high-level visual tasks, such as image classification, tracking, and object detection. Therefore, image dehazing is a practical problem. One example is the single image dehazing task, where the aim is to recover a clean image from a single contaminated image.

To this end, traditional methods \cite{tan2008visibility,fattal2008single,tarel2009fast,he2010single,meng2013efficient,tang2014investigating,zhu2015fast,berman2016non} rely on the atmospheric scattering model \cite{mccartney1976optics,narasimhan2000chromatic,narasimhan2002vision}, which provides a simple approximation of the haze effect by assuming that:
\begin{equation}
    \label{Eq: ASM}
    \mathbf{I}(x) = \mathbf{J}(x)t(x) + A(1 - t(x)),
\end{equation}
\noindent where $\mathbf{I}$ is a hazy image composed of its corresponding haze-free image $\mathbf{J}$ and the global atmospheric light $A$. Moreover, we have the transmission map $t(x) = e^{-\beta d(x)}$, with $\beta$ and $d(x)$ being the atmospheric scattering parameter and the scene depth, respectively. To recover the haze-free image $\mathbf{J}$, these methods estimate $A$ and $t(x)$ from the hazy image $\mathbf{I}$ by means of handcrafted priors. However, although these traditional methods perform well in some cases, their strong dependency on priors may lead to poor estimation results on the transmission map when these priors do not hold. Therefore, recent research has focused on deep learning-based methods to avoid the need to search for robust handcrafted priors. By relying upon the strong learning capacity, these methods can directly learn the mapping from hazy to haze-free domains from existing datasets.

Deep learning-based methods have achieved breakthroughs in terms of image quality and fidelity; however they are subject to demanding training times and large-scale paired datasets consisting of diverse data. Moreover, the collection of paired data for image dehazing is expensive and time-consuming. In addition, existing datasets are typically biased to specific environments. Consequently, these methods suffer from deteriorated performance when confronting domain shifts due to variations in position, weather conditions, and haze density. Hence, it is imperative to improve the data efficiency of these deep learning-based methods to overcome domain shifts in practice.

\begin{figure*}[!ht]
    \centering
    \includegraphics[width=0.9\textwidth]{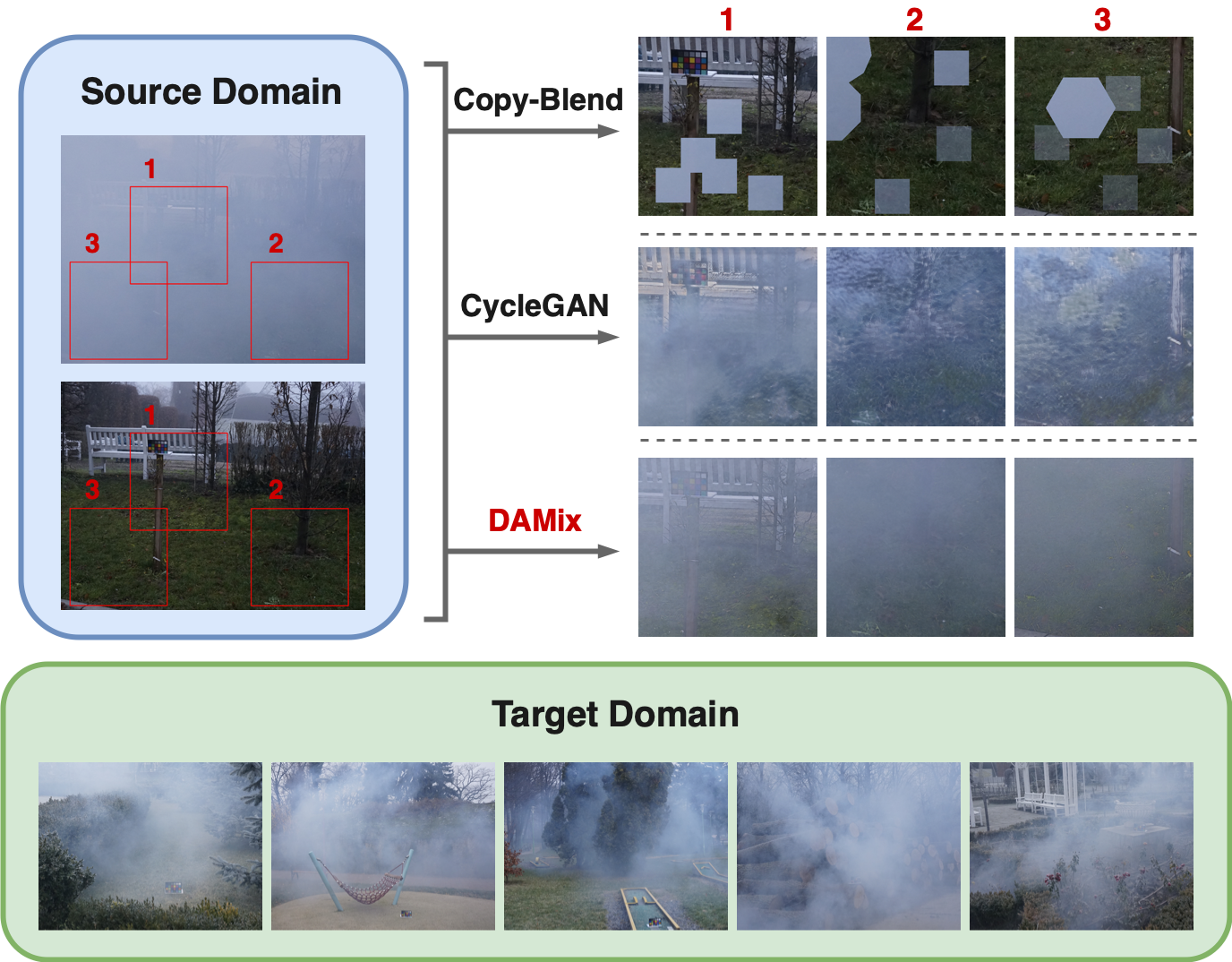}
    \caption{Samples generated by Copy-Blend \cite{shyam2021evaluating}, CycleGAN \cite{CycleGAN2017}, and \textbf{DAMix}. Copy-Blend and CycleGAN both yield artifacts or unnatural appearance. In contrast, DAMix generates samples with the most visually pleasing appearance and also features a similar haze distribution of the hazy images in the target domain.}
    \label{fig: mainfig}
\end{figure*}

The community has conducted several studies to address these issues. Two key examples are domain adaptation (DA) and domain generalization (DG). DA utilizes a labeled source domain and unlabeled target domain to develop a model that performs well on the target domain. The goal of DG is to learn a model from one or multiple source domains that may generalize well on unseen target domains. Both tasks aim to mitigate the domain shift issue, and the key difference between them is the access to unlabeled data in the target domain. Since unlabeled data are generally available in single image dehazing, we will focus more on DA in this paper.

Data augmentation is a simple but effective method that is widely used to improve data efficiency and generalized performance. Prior work \cite{shao2020domain,Shyam_Yoon_Kim_2021} manipulates hazy images to obtain more diverse training pairs and further bypasses domain shifts. \cite{shao2020domain} employs CycleGAN \cite{CycleGAN2017} to translate images between synthetic and real domains. By training with translated samples, the dehazing network can achieve better results on the real domain. \cite{Shyam_Yoon_Kim_2021} employs Copy-Blend augmentation \cite{shyam2021evaluating} to simulate nonhomogeneous haze distributions. Although these methods are effective in certain settings, they suffer from the following limitations. (1) Copy-Blend is not sufficiently general since it is an ad hoc method for nonhomogeneous haze. (2) Training a CycleGAN model consumes considerable amounts of time and GPU resources. (3) As shown in Fig. \ref{fig: mainfig}, Copy-Blend and CycleGAN produce samples with artifacts. Consequently, learning with these samples leads to unstable results. (4) The generated samples of these methods lack interpretability.

Given the above issues, we aim to design a data augmentation method to mitigate domain shifts in single image dehazing. Since previous work does not leverage the mechanism of the haze formulation (\ref{Eq: ASM}), we consider the prior work to be more general-purpose and not designed specifically for dehazing: we observe that mixup \cite{zhang2017mixup} is well-suited for this task. Mixing the hazy image with its corresponding ground-truth image or the global atmospheric light can make the haze of the mixup-ed output either thinner or thicker. Moreover, the mixup-ed output is guaranteed to comply with the atmospheric scattering model (\ref{Eq: ASM}). Mixup has the potential to generate complex and novel training data for free. Therefore, we propose density-aware mixup data augmentation (DAMix). DAMix aims to reduce the performance degradation caused by domain shifts while simultaneously improving the data efficiency. As shown in Fig. \ref{fig: mainfig}, DAMix generates synthetic hazy samples featuring the haze distribution of the target domain. These samples are generated by mixing up the hazy image with its ground truth or global atmospheric light by means of a density-aware combination ratio. After learning from these DAMix-ed samples with diverse haze distributions, deep learning-based methods can generalize better when encountering domain shifts. In summary, our contributions offer a number of advantages over the existing methods:

\begin{itemize}
    \item \textbf{Interpretability}\quad We formulate DAMix as a constrained optimization problem. Then, we propose an efficient algorithm employing the exact histogram specification \cite{coltuc2006exact}. Eventually, leveraging the mixup operation \cite{zhang2017mixup}, DAMix-ed samples are proved to comply with the atmospheric scattering model (\ref{Eq: ASM}). Consequently, DAMix-ed samples are more interpretable and visually pleasing compared to those in previous work (see Fig. \ref{fig: mainfig}).
    \item \textbf{Domain Shift}\quad DAMix can generate samples with an unlimited haze distribution, thereby mitigating the haze density gap between domains. In addition, DAMix can further improve the performance of various deep learning-based methods when confronting domain shifts.
    \item \textbf{Data Efficiency}\quad DAMix improves the data efficiency of state-of-the-art image dehazing models. In particular, when there is a domain shift between the source and target domains, the model trained with DAMix using only half of the source dataset can adapt as well as that trained with the whole source dataset but without DAMix.
    \item \textbf{Easy-to-use}\quad Since DAMix does not require training, it consumes less computational resources. Moreover, DAMix operates only on the data level; we can directly utilize it with any deep learning-based method to improve performance.
\end{itemize}

\section{Related Work}
Single image dehazing aims to recover a haze-free image from a corresponding hazy image. The existing dehazing methods can be classified into prior-based methods and deep learning-based methods.
    \subsection{Prior-based Methods}
        Most prior-based methods rely on the atmospheric scattering model (\ref{Eq: ASM}) and employ strong priors or assumptions to restore the transmission map and the global atmospheric light from a hazy image. Representative works utilizing prior-based methods include \cite{he2010single,zhu2015fast,berman2016non}. \cite{he2010single} introduces the dark channel prior as an additional constraint to estimate haze-free images. The dark channel prior is based on the observation that in haze-free patches, at least one of the color channels has some pixels whose intensity is very low or close to zero. \cite{zhu2015fast} proposes the color attenuation prior, which assumes that the scene depth is positively correlated with the difference between the brightness and the saturation. \cite{berman2016non} relies on the assumption that a few distinct colors can effectively represent the colors of a haze-free image. Due to the haze effect, these varying distances translate to different transmission coefficients; therefore, each color cluster in the haze-free image becomes a haze-line in RGB space. Nevertheless, although these methods show good performance under certain circumstances, their strong dependency on priors and the atmospheric scattering model may degrade the performance when confronted with complex realistic scenes not contained within the priors or the atmospheric scattering model.
    \subsection{Deep Learning-based Methods}
        Early works follow the atmospheric scattering model (\ref{Eq: ASM}) by directly estimating the global atmospheric light and the transmission map from the corresponding hazy image. For example, \cite{li2017aod} utilizes a simple CNN-based architecture to predict the transmission map and the global atmospheric light. However, the performance of these works relies strongly on estimating the transmission map and the global atmospheric light. Therefore, recent works have focused on the end-to-end prediction of haze-free images from hazy images. \cite{liu2019griddehazenet} proposes an autoencoder endowed with a multiscale and channelwise attention mechanism. To focus more on regions with thick haze, \cite{qin2020ffa} proposes a network utilizing a feature attention module that combines channel attention and pixel attention mechanisms. \cite{dong2020multi} incorporates the SOS boosting strategy \cite{romano2015boosting} and the back-projection technique for image dehazing. \cite{hong2020distilling} proposes a knowledge distillation network by learning additional information from an intermediate representation of the teacher network. \cite{wu2021contrastive} applies the method of contrastive learning to image dehazing by pulling the anchor (model output) close to positive points (ground-truth) and pushing the anchor far from negative points (hazy input) in the feature space. Unfortunately, although these works report admirable performance, they suffer from substantial performance drops when confronting domain shifts; recently, several works have attempted to address this issue. \cite{shao2020domain} indicates that a model trained on synthetic hazy images cannot exhibit sufficient generalization performance on real hazy images and therefore introduces a domain adaptation paradigm with an image translation module \cite{CycleGAN2017} to bridge the gap between the synthetic and real domains. In addition, \cite{Shyam_Yoon_Kim_2021} applies frequency information to adversarial training with the aggregated dataset in combination with Copy-Blend \cite{shyam2021evaluating} to achieve domain-invariant performance.

\section{Method}
    This section shows the density-aware mixup data augmentation method (DAMix) for training the image dehazing network. We first present the notation needed for this section. Then, we define the haze density applied in DAMix. Next, the detailed procedure of DAMix is shown and described as a constrained optimization problem. Then, we introduce an efficient haze density alignment algorithm that provides an approximate solution for the optimization problem. Finally, we exhibit how we sample the haze density target to achieve randomization in DAMix.
    \subsection{Notations}
    Throughout the paper, let $\Omega$ be the image domain, typically $\Omega=\{1, ..., H\}\times\{1, ..., W\}$ for discrete cases, where $H$ and $W$ denote the height and width, respectively. We denote an image by $I: \Omega \rightarrow \mathcal{M}^{c}$, where $\mathcal{M}^{c}=\{0, ..., n-1\}^{c}$ is the range of the pixel value. For 8-bit precision, we generally have $n=256$, for gray-level images we have $c=1$ and for color images $c=3$.
    \subsection{On the Estimation of Haze Density}
    \label{sec: haze density estimation}
\begin{figure}[!ht]
        \centering
        \includegraphics[width=0.95\linewidth]{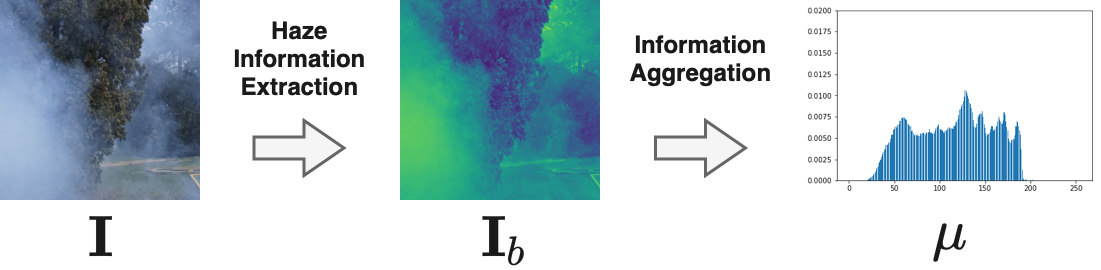}
        \caption{The haze density estimation procedure.}
        \label{fig: HazeEstimation}
    \end{figure}

    As shown in Fig. \ref{fig: HazeEstimation}, our haze density estimation can be separated into two steps: haze information extraction and information aggregation.
    
    \noindent \textbf{Haze Information Extraction:} According to the color attenuation prior \cite{zhu2015fast}, haze density is positively correlated with brightness. We thus leverage the brightness to extract the haze information from the hazy image. Given a hazy image $\mathbf{I}$, we first obtain its corresponding one-channel brightness image $\mathbf{I}_{b}: \Omega \rightarrow \mathcal{M}$.
    
    \noindent \textbf{Information Aggregation:} Due to the complex haze distribution, it is not appropriate to apply a scalar-valued function as the estimate of haze density. We employ the probability measure $\mu$ over $\mathcal{M}$ as the representation of the haze density to preserve the characteristics of the haze. We denote $\Sigma \subset 2^{\mathcal{M}}$ as a Borel $\sigma$-algebra over $\mathcal{M}$. Then, we consider the probability measure $\mu: \Sigma \rightarrow [0,1]$, which records the value $\mathbf{I}_{b}$ takes:
    \begin{equation}
        \label{Eq: mu_I}
        \mu=(\sum_{x \in \Omega}\delta_{\mathbf{I}_{b}(x)})/|\Omega|,
    \end{equation}
    where $\delta_{v}$ is the Dirac delta function at position $v$. It follows that for any $m \in \Sigma$ we have $\mu(m)=(\sum_{x \in \Omega}\mathbbm{1}_{\{\mathbf{I}_{b}(x)\in m\}})/|\Omega|$, which indicates the proportion of area where $\mathbf{I}_{b}$ takes a value in $m \subset \mathcal{M}$. Therefore, the probability measure $\mu$ over $\mathcal{M}$ can accurately capture the distribution of the haze density within $\mathbf{I}$, regardless of the homogeneity of the haze.
    
    \subsection{DAMix Formulation}
    \label{sec: DAMix_form}
    Given a hazy image and its corresponding haze-free image $\mathbf{I},\mathbf{J}: \Omega \rightarrow \mathcal{M}^{3}$, the atmospheric scattering model (\ref{Eq: ASM}) provides a simple approximation of the haze effect by assuming that $\mathbf{I}$ is composed of $\mathbf{J}$ and the global atmospheric light $A \in \mathcal{M}^{3}$ with the transmission map $t$ acting as the combination ratio between $\mathbf{J}$ and $A$. The haze effect is highly correlated with $t$. A simple data augmentation method can thus be derived by increasing the proportion of $\mathbf{J}$ or $A$ to manipulate $t$:
    \begin{equation}
        \label{Eq: EasyDAMix}
        \hat{\mathbf{I}}=
        \begin{cases}
            \lambda \mathbf{I} + (1-\lambda) \mathbf{J} \quad\text{for thinner haze}\\
            \lambda \mathbf{I} + (1-\lambda) \mathbf{A} \quad\text{for thicker haze}
        \end{cases},
    \end{equation}
    where $\lambda \in [0,1]$. Moreover, $\mathbf{A}: \Omega \rightarrow \mathcal{M}^{3}$ is an image such that $\forall x \in \Omega,\; \mathbf{A}(x)=A$. Typically, $A$ can be estimated via prior-based methods. In this paper, we apply the famous dark channel prior \cite{he2010single} to estimate $A$. 
    
    Although Eq. (\ref{Eq: EasyDAMix}) provides a straightforward image transformation to modify the haze density, it restricts the haze of the entire hazy image to be either thinner or thicker. To make the transformation more general, we increase the degrees of freedom and transform Eq. (\ref{Eq: EasyDAMix}) as follows:
    \begin{equation}
        \label{Eq: DAMix}
        \hat{\mathbf{I}} = (1-\alpha-\beta) \odot \mathbf{I} + \alpha \odot \mathbf{J} + \beta \odot \mathbf{A},
    \end{equation}
    where $\alpha,\beta \in \mathbb{R}_{+}^{H\times W}$ and $\alpha+\beta \leq 1$. $\odot$ denotes the Hadamard product.
    
    Let us imagine a scenario in which we are given a haze density target $\mu^{t}$. We first derive the haze density of $\hat{\mathbf{I}}$ as $\hat{\mu}$ according to the estimation mentioned in Sec. \ref{sec: haze density estimation}. Then, we try to obtain feasible $\alpha$ and $\beta$ values such that the distance between $\hat{\mu}$ and $\mu^{t}$ is minimized, which can be formulated as follows:
    \begin{equation}
        \label{Eq: opt}
        \alpha^{\star}, \beta^{\star} = \arg\min_{\alpha, \beta}\mathcal{E}(\alpha, \beta),\;\textbf{where}\; \mathcal{E}(\alpha, \beta)\coloneqq W_{p}(\hat{\mu},\mu^{t}),
    \end{equation}
    and $W_{p}(\cdot, \cdot)$ denotes the $p$-Wasserstein distance \cite{villani2009optimal}. By solving the constrained optimization problem in Eq. (\ref{Eq: opt}), we can obtain a synthetic hazy image $\hat{\mathbf{I}}$ that has a haze distribution similar to $\mu^{t}$.

    \subsection{Haze Density Alignment}
    The energy $\mathcal{E}(\alpha, \beta)$ in Eq. (\ref{Eq: opt}) can be solved by projected gradient descent in general (more details are shown in the supplementary material). Although projected gradient descent can achieve satisfactory results, the optimization iterations result in excessive computational and time complexity. Therefore, this iterative method should not be employed during network training.
    
    To resolve the high computational complexity, we propose an efficient haze density alignment algorithm to find approximate solutions to Eq. (\ref{Eq: opt}). The energy $\mathcal{E}(\alpha, \beta)$ is minimized when $\hat{\mu}$ and $\mu^{t}$ are exactly matched. For clarity, we represent the haze density target $\mu^{t}$ as a normalized histogram $h^{t} \in \mathbb{R}^{n}_{+}$ such that:
    \begin{equation}
        h^{t}_{i}=\mu^{t}(\{i\}),
    \end{equation}
    where $i \in \mathcal{M}$. To exactly match with $h^{t}$, a strict ordering on image pixels of $\mathbf{I}_b$ is required. We thus follow the ordering of the exact histogram specification \cite{coltuc2006exact} which employs the local mean to achieve a strict ordering almost everywhere. It follows that the ordered pixels can be split into $n$ groups such that group $j$ has $h^{t}_{j}$ pixels. In this manner, we can obtain a one-channel prototype $\mathbf{I}_{p}: \Omega \rightarrow \mathcal{M}$ that features an identical histogram to $h^{t}$. Then, we can utilize Eq. (\ref{Eq: DAMix}) to obtain feasible $\alpha$ and $\beta$ values such that $\mathbf{I}_{p}=(1-\alpha-\beta) \odot \mathbf{I}_{b} + \alpha \odot \mathbf{J}_{b} + \beta \odot \mathbf{A}_{b}$. Since haze increases the brightness in general \cite{zhu2015fast}, we assume that $\mathbf{A}_{b} \geq \mathbf{I}_{b} \geq \mathbf{J}_b$. We further constrain $\alpha \odot \beta = 0_{H,W}$ to make the solution of $\alpha, \beta$ deterministic. The feasible solution can thus be obtained by projecting the solution onto the feasible set, which can be represented as follows:
    \begin{equation}
    \begin{aligned}
        &\alpha^{\star}(x) = \min(\frac{(\mathbf{I}_{b}(x)-\mathbf{I}_{p}(x))\cdot \mathbbm{1}_{\{\mathbf{I}_{b}(x) \geq \mathbf{I}_{p}(x)\}}}{\mathbf{I}_{b}(x)-\mathbf{J}_{b}(x)}, 1)\\
        &\beta^{\star}(x) = \min(\frac{(\mathbf{I}_{p}(x)-\mathbf{I}_{b}(x))\cdot \mathbbm{1}_{\{\mathbf{I}_{b}(x) < \mathbf{I}_{p}(x)\}}}{\mathbf{A}_{b}(x)-\mathbf{I}_{b}(x)}, 1)
    \end{aligned}
    \end{equation}
    Following Eq. (\ref{Eq: DAMix}), the DAMix-ed image $\hat{\mathbf{I}}$ can be formulated as follows:
    \begin{equation}
        \hat{\mathbf{I}}=(1-\alpha^{*}-\beta^{*}) \odot \mathbf{I} + \alpha^{*} \odot \mathbf{J} + \beta^{*} \odot \mathbf{A}.
    \end{equation}
    
    Unlike existing methods that often yield artifacts in their synthetic samples, DAMix is a robust algorithm that generates natural samples because $\hat{\mathbf{I}}$ complies with the atmospheric scattering model (\ref{Eq: ASM}):
    \begin{equation}
        \begin{aligned}
        \label{eq:DAMix_proof}
            \hat{\mathbf{I}}(x)
            &= \textbf{J}(x)[(1-t(x))\alpha^{\star}(x)+t(x)(1-\beta^{\star}(x))]\\
            &+ A[1-(1-t(x))\alpha^{\star}(x)-t(x)(1-\beta^{\star}(x))]\\
            &= \textbf{J}(x)\hat{t}(x) + A(1 - \hat{t}(x)).
        \end{aligned}
    \end{equation}
    Evidently, DAMix generates samples by modifying the transmission map such that the new transmission map $\hat{t}(x)=(1-t(x))\alpha^{\star}(x)+t(x)(1-\beta^{\star}(x))$; thus, DAMix-ed samples can retain their natural appearance despite the translation.
    
    \begin{figure}[!ht]
        \centering
        \includegraphics[width=0.95\linewidth]{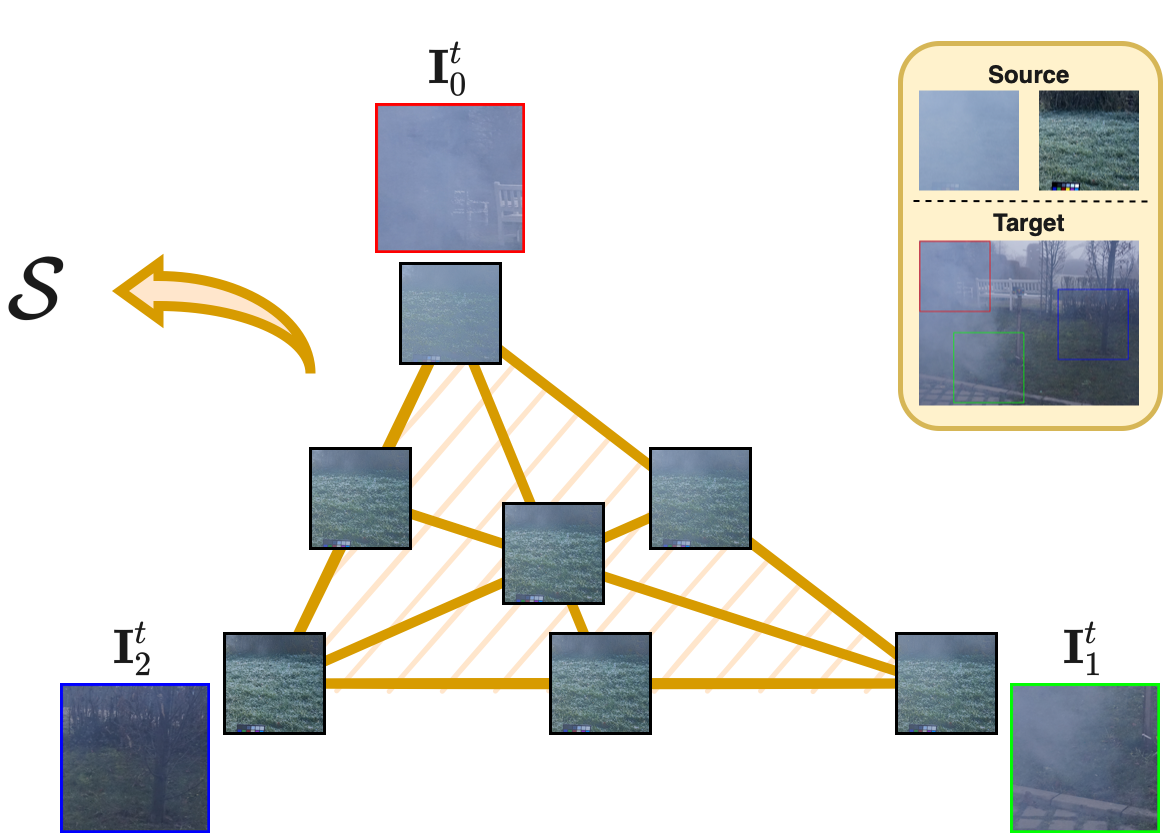}
        \caption{This figure exhibits the set $\mathcal{S}$. The vertices of the triangle represent the positions of $\mu^{t}_{i}$ in the Wasserstein space. Each DAMix-ed image (\textit{black box bounded}) is placed at the position of the corresponding sampled $\mu^{t}$ it uses.}
        \label{fig: mu_t}
    \end{figure}
    \subsection{Acquisition of Haze Density Target}
    We have shown the role of the haze density target $\mu^{t}$ in Eq. (\ref{Eq: opt}); here, we introduce how $\mu^{t}$ is derived. Imagine a scenario in which we have $K$ hazy images $\mathbf{I}^{t}_{i}$, where $i\in\{0,...,K-1\}$ in the target domain. We first obtain the corresponding haze density of each hazy image $\mu^{t}_{i}$ according to Sec. \ref{sec: haze density estimation}. Then, during training, we sample $\mu^{t}$ from the set $\mathcal{S}$ that includes all interpolations of $\mu^{t}_{i}$ in the Wasserstein space. According to (7.7) in \cite{peyre2019computational}, $\mathcal{S}$ can be formulated as follows:
    \begin{equation}
        \mathcal{S} = \{\mu|\mathcal{C}^{-1}_{\mu}=\Sigma_{i=0}^{K-1}\theta_{i}\mathcal{C}^{-1}_{\mu^{t}_{i}},\Sigma_{i=0}^{K-1}\theta_{i}=1,\theta_{i} \geq 0\},
    \end{equation}
    where $\mathcal{C}_{\mu}^{-1}$ is the pseudoinverse of the cumulative distribution function of $\mu$, namely, the generalized quantile function of $\mu$. We provide a toy example with $K=3$ in Fig. \ref{fig: mu_t}, which shows the samples generated by DAMix under different interpolations of $\mu^{t}_{i}$ in the Wasserstein space. The similarity of the haze distribution of DAMix-ed samples to that of $\mathbf{I}^{t}_{i}$ is correlated with the ratio $\theta_{i}$.   

\section{Experiments}
\subsection{Experimental Settings}
\subsubsection{Datasets and Evaluation Metrics.}
To evaluate the effectiveness of DAMix, we select three datasets with different haze densities: Dense-Haze \cite{Dense-Haze_2019}, NH-Haze \cite{NH-Haze_2020}, O-Haze \cite{O-HAZE_2018}. These datasets were introduced in the NTIRE challenge \cite{ancuti2018ntire,NTIRE_Dehazing_2019,NTIRE_Dehazing_2020}. Since the haze in these datasets is produced by professional haze machines, they are more challenging than synthetic datasets. Furthermore, there are haze density differences among them. 
\begin{table*}[!ht]
    \renewcommand\arraystretch{1.5} 
    \small
    \centering
    \resizebox{\textwidth}{!}{
    \begin{tabular}{ |c||cc|cc|cc| }
        \hline
        \multirow{2}{*}{\makecell[c]{Settings}} &
        \multicolumn{2}{c|}{\textbf{GridDehazeNet} \cite{liu2019griddehazenet}} &
        \multicolumn{2}{c|}{\textbf{FFA-Net} \cite{qin2020ffa}} &
        \multicolumn{2}{c|}{\textbf{MSBDN} \cite{dong2020multi}}\\
        & w/o DAMix & w/ DAMix & w/o DAMix & w/ DAMix & w/o DAMix & w/ DAMix\\
        \hline
        $D \rightarrow N$  &
        15.78/0.593 &
        \textbf{16.53/0.608} &
        16.60/\textbf{0.612} &
        \textbf{16.86}/0.608 &
        15.87/0.595 &
        \textbf{17.26/0.630}\\
        $D \rightarrow O$ &
        16.11/0.579 &
        \textbf{20.22/0.686} &
        16.91/0.623 &
        \textbf{19.45/0.647} &
        18.39/0.673 &
        \textbf{21.46/0.693}\\
        \hline
        $N \rightarrow D$  &
        14.48/0.481 &
        \textbf{14.66/0.510} &
        11.71/\textbf{0.379} &
        \textbf{12.89}/0.368 &
        14.74/0.506 &
        \textbf{14.99/0.511}\\
        $N \rightarrow O$ &
        18.38/0.623 &
        \textbf{18.41/0.643} &
        17.25/0.547 &
        \textbf{18.45/0.580} &
        18.73/\textbf{0.649} &
        \textbf{19.81}/0.637\\
        \hline
        $O \rightarrow D$  &
        12.92/0.502 &
        \textbf{13.03/0.510} &
        13.06/\textbf{0.518} &
        \textbf{13.35}/0.510 &
        13.69/\textbf{0.539} &
        \textbf{14.07}/0.529\\
        $O \rightarrow N$ &
        16.21/0.556 &
        \textbf{16.38/0.575} &
        16.91/0.588 &
        \textbf{17.10/0.610} &
        16.42/0.571 &
        \textbf{16.76/0.581}\\
        \hline
    \end{tabular}}
    \caption{Quantitative improvements in the domain adaptation performance by applying DAMix. We adopt \textbf{PSNR/SSIM} as evaluation metrics. \textbf{Boldface} indicates the best results for each setting.}
    \label{tab: Adaptation}
\end{table*}
They can be sorted according to haze density in descending order as Dense-Haze, NH-Haze, and O-Haze. In addition, we use the Outdoor Training Set (OTS) and Unannotated Realistic Hazy Images (URHI) of RESIDE \cite{li2019benchmarking} in Sec. \ref{sec: Real} to verify the adaptivity of DAMix from synthetic datasets to the real-world samples. More properties of these datasets are provided in the supplementary material. Additionally, we utilize PSNR and SSIM as evaluation metrics to compare the performance of different methods.
\subsubsection{Architecture.}
We use recent open-source deep learning-based methods (GridDehazeNet \cite{liu2019griddehazenet}, FFA-Net \cite{qin2020ffa}, MSBDN \cite{dong2020multi}) to evaluate the robustness and effectiveness of DAMix on domain adaptation (Sec. \ref{sec: DA}). In the remaining sections, we apply GridDehazeNet \cite{liu2019griddehazenet} to conduct experiments due to its lightweight design and computational efficiency.

\subsection{DAMix Helps Domain Adaptation}
\label{sec: DA}
We conduct thorough experiments on all the combinations of the three chosen datasets (Dense-Haze, NH-Haze, and O-Haze) and present quantitative comparisons and qualitative comparisons.

\subsubsection{Quantitative Comparisons.}
For simplicity, we denote Dense-Haze \cite{Dense-Haze_2019}, NH-Haze \cite{NH-Haze_2020}, and O-Haze \cite{O-HAZE_2018} as  $D$, $N$, and $O$, respectively. Moreover, $A \rightarrow B$ represents the setting where $A$ is the source domain and $B$ is the target domain. As shown in Tab. \ref{tab: Adaptation}, all the models notably improve in terms of PSNR in almost all settings when DAMix is applied in the training phase. Although there are a few settings with a minor drop in SSIM, it appears to be an acceptable trade-off due to the corresponding greater increase in PSNR. Additionally, as the difference in the haze density between the source and target domains increases, models trained with DAMix show greater improvement in terms of PSNR/SSIM. Specifically, DAMix helps more in the $D \rightarrow O$ setting than the $D \rightarrow N$ setting in general. Moreover, settings that adapt to the target domain with thinner haze can achieve a better improvement because $\mathbf{J}$ is given, but $A$ is obtained via estimation. Consequently, DAMix is subject to the precision of $A$ when adapting to domains with thicker haze. As we can see, DAMix achieves average performance gains equaling a PSNR of only 0.26 dB in the $O \rightarrow D$ setting, but DAMix improves the baseline by an average of 3.24 dB PSNR in the $D \rightarrow O$ setting.

\subsubsection{Qualitative Comparisons.}
\begin{figure*}[!ht]
    \centering
    \captionsetup{justification=centering,font={small}}
    \begin{subfigure}[t]{.24\textwidth}
        \includegraphics[width=.9\linewidth]{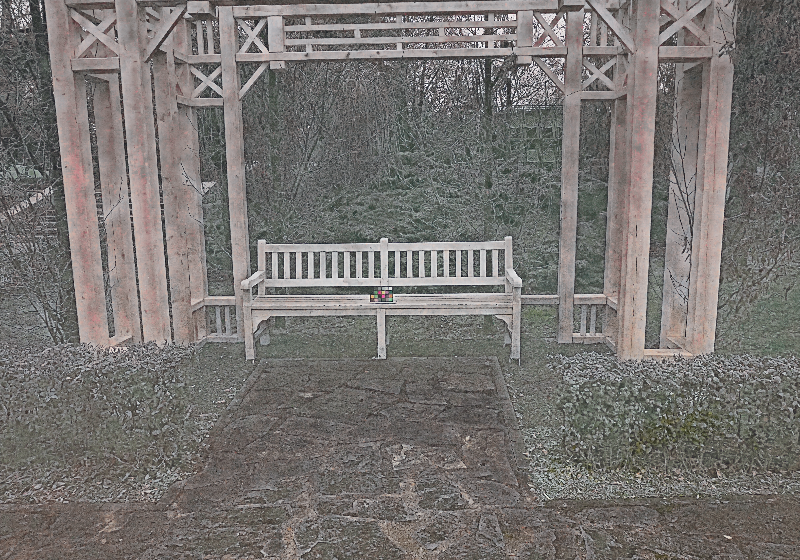}
        \caption{GridDehazeNet\\(w/o DAMix)}
    \end{subfigure}
    \begin{subfigure}[t]{.24\textwidth}
        \includegraphics[width=.9\linewidth]{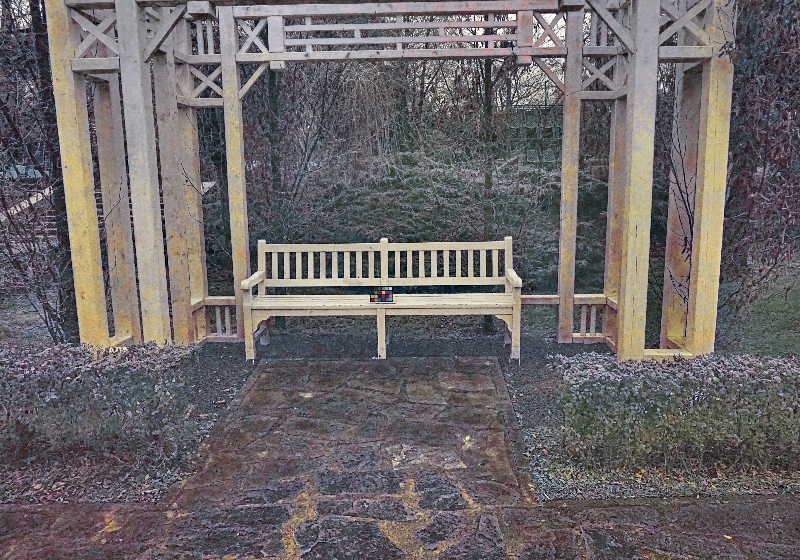}
        \caption{FFA-Net\\(w/o DAMix)}
    \end{subfigure}
    \begin{subfigure}[t]{.24\textwidth}
        \includegraphics[width=.9\linewidth]{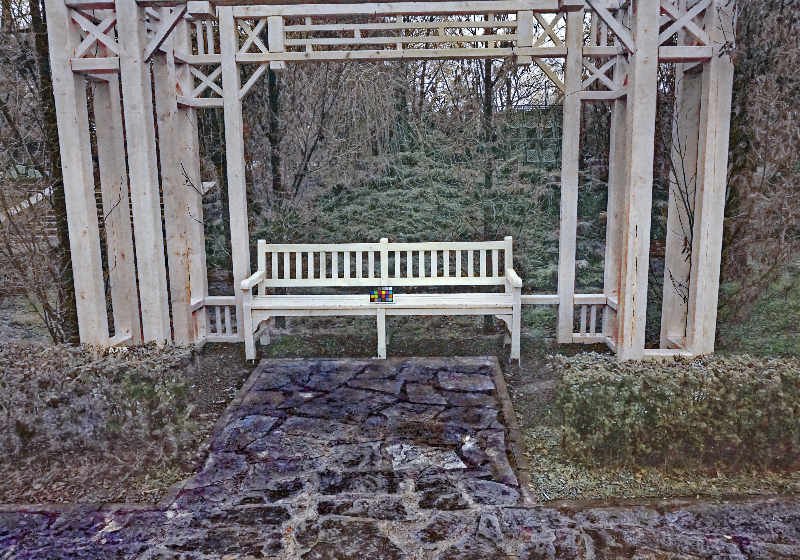}
        \caption{MSBDN\\(w/o DAMix)}
    \end{subfigure}
    \begin{subfigure}[t]{.24\textwidth}
        \includegraphics[width=.9\linewidth]{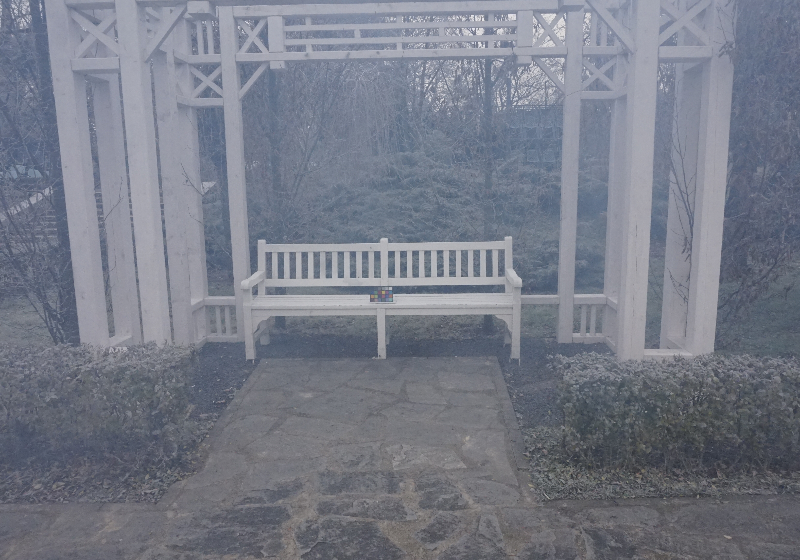}
        \caption{Hazy}
    \end{subfigure} \\
    \begin{subfigure}[t]{.24\textwidth}
        \includegraphics[width=.9\linewidth]{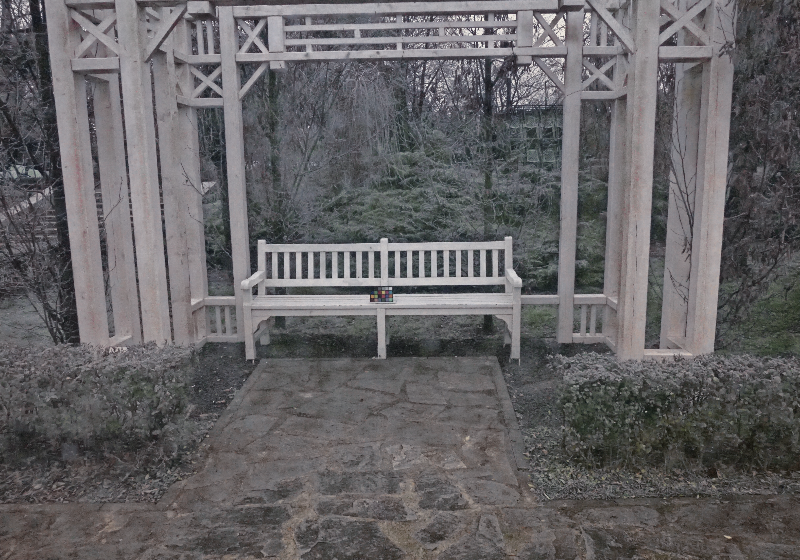}
        \caption{GridDehazeNet\\(w/ DAMix)}
    \end{subfigure}
    \begin{subfigure}[t]{.24\textwidth}
        \includegraphics[width=.9\linewidth]{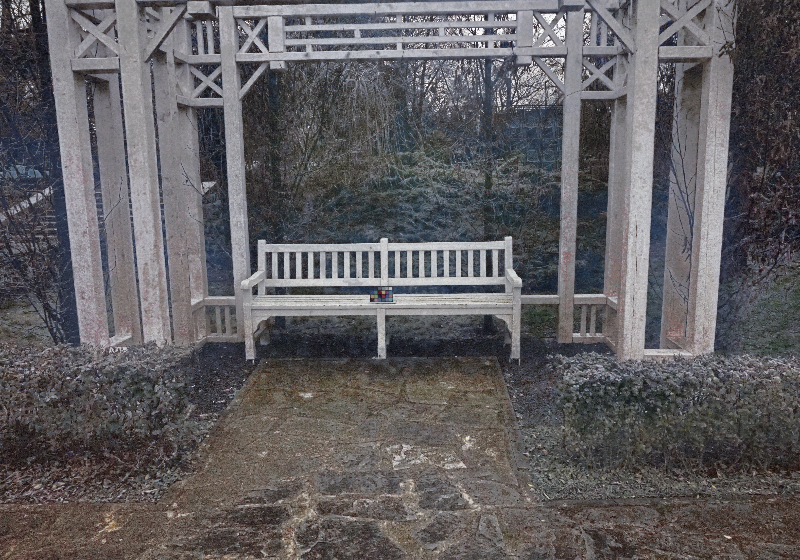}
        \caption{FFA-Net\\(w/ DAMix)}
    \end{subfigure}
    \begin{subfigure}[t]{.24\textwidth}
        \includegraphics[width=.9\linewidth]{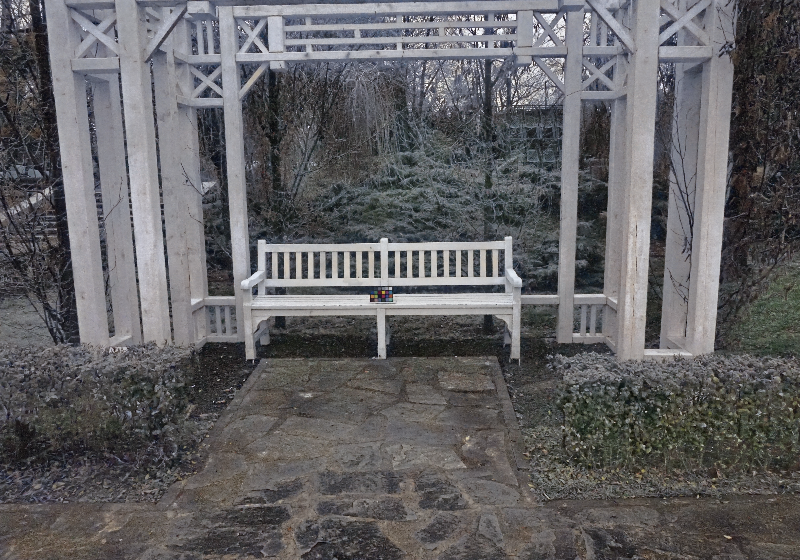}
        \caption{MSBDN\\(w/ DAMix)}
    \end{subfigure}
    \begin{subfigure}[t]{.24\textwidth}
        \includegraphics[width=.9\linewidth]{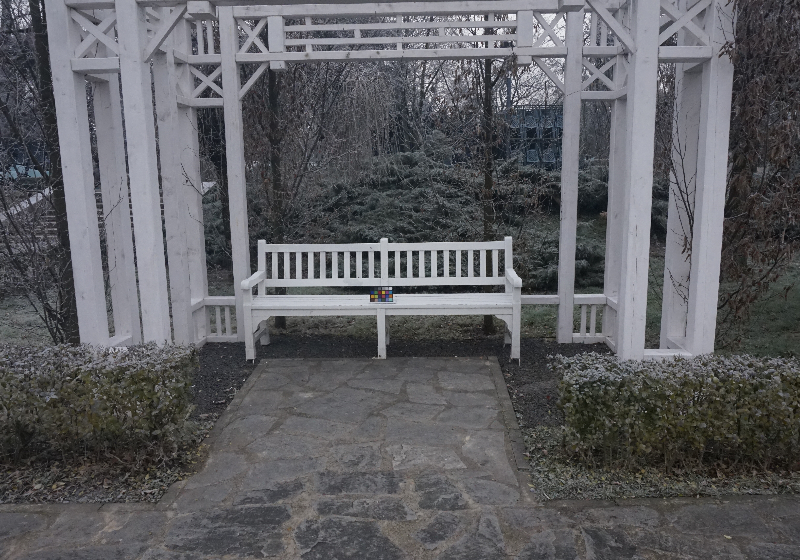}
        \caption{GT}
    \end{subfigure}
    \captionsetup{justification=justified,font={normalsize}}
    \caption{Qualitative comparisons of the models trained on Dense-Haze and tested on O-Haze.}
    \label{fig: Adaptation}
\end{figure*}

In Fig. \ref{fig: Adaptation}, we provide examples exhibiting the effectiveness of DAMix when models confront domain shifts. The models trained without DAMix suffer from a color shift and overenhancement on the pillars and the floor because the models trained without DAMix only learn to recover clear images from highly contaminated images. However, with the help of DAMix, these models can learn from samples that feature similar haze density to that of the target domain during training. These models can then recover more precise and realistic colors when tested on the target domain.

\subsection{DAMix Helps Data Efficiency}
\label{sec: DE}
In this section, we show that DAMix can improve data efficiency under different perspectives. We design experiments using only partitions of the dataset to train the network, simulating data shortage in practice. In addition, we apply existing methods \cite{shyam2021evaluating,CycleGAN2017} to serve as strong baselines. The experiments are conducted under three different perspectives simulating conditions that exist in reality. 

\subsubsection{Without Domain Shift.}
\begin{figure}[!ht]
    \centering
    \includegraphics[width=.9\linewidth]{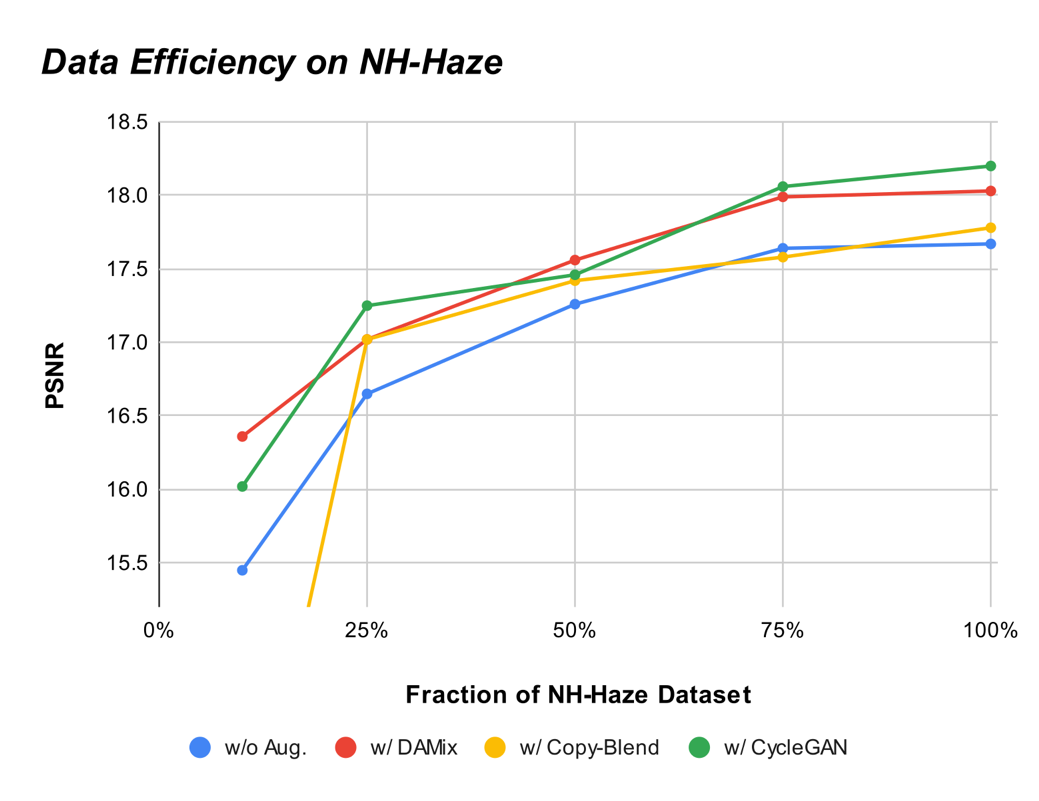}
    \caption{Data efficiency on the NH-Haze \cite{NH-Haze_2020} benchmark.}
    \label{fig: DE_N}
\end{figure}

Here, we present an experiment with no or few domain shifts between the source and target domain. Since Copy-Blend \cite{shyam2021evaluating} is designed to simulate the nonhomogeneous haze, we select NH-Haze \cite{NH-Haze_2020} to verify the effectiveness. We use the train and test sets of NH-Haze as the source and target domains, respectively. In addition, we use only partitions of the training data ($10\%,25\%,50\%,75\%,100\%$) to simulate the shortage of paired data in reality. Fig. \ref{fig: DE_N} shows that DAMix and CycleGAN \cite{CycleGAN2017}
are always helpful across all fractions of NH-Haze \cite{NH-Haze_2020}, whereas Copy-Blend fails to improve in the low data regime ($10\%$). In contrast, DAMix is most helpful in the low data regime, yielding a 0.91 dB PSNR improvement on top of the network trained without any of the three data augmentation methods. We attribute this improvement to the diversity of DAMix, which offers the network more synthetic hazy images with different haze distributions. Although CycleGAN shows slightly better improvements in a few settings ($25\%,75\%,100\%$), it requires considerable amounts of time and GPU resources (approximately two days with one NVIDIA V100) to pretrain the image translation model. Hence, we consider DAMix the superior choice due to the better time and performance trade-off.

\subsubsection{Domain Adaptation.}
\begin{figure*}[!ht]
    \centering
    \includegraphics[width=.95\textwidth]{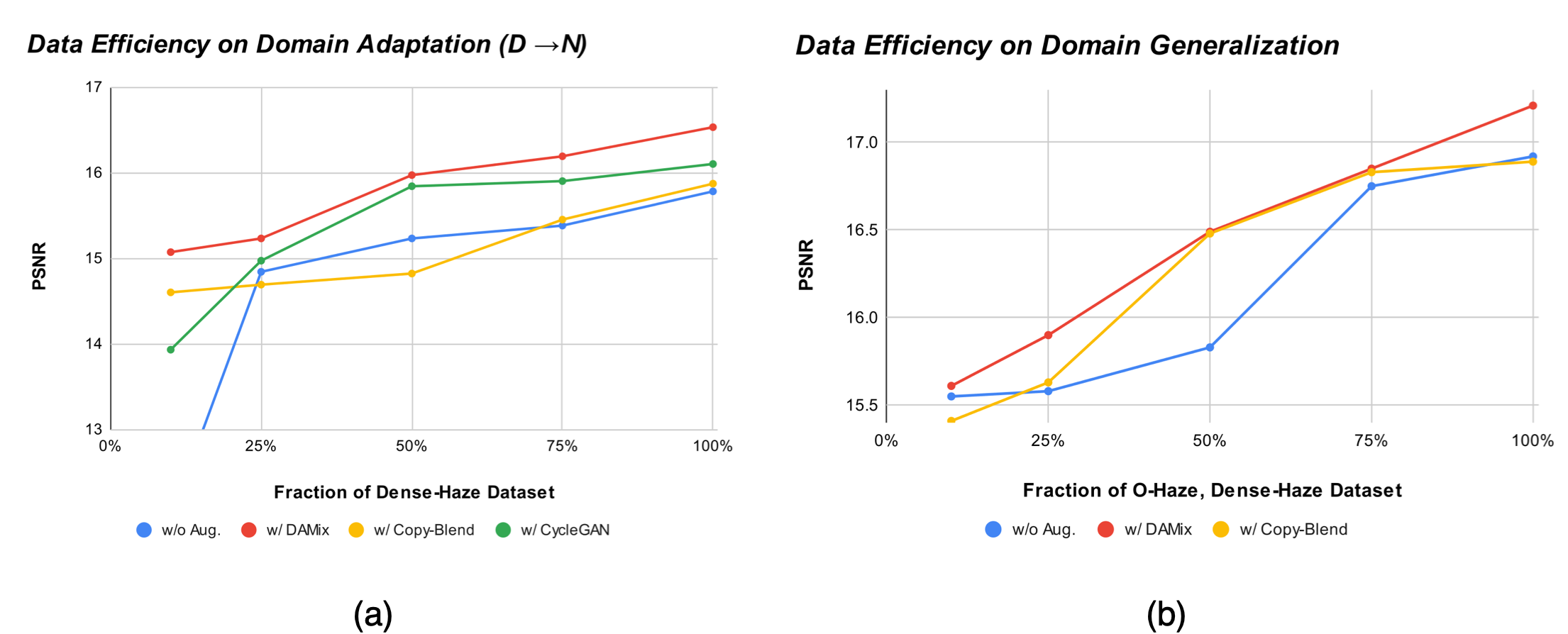}
    \caption{Data efficiency when confronting domain shifts.}
    \label{fig: DE_DS}
\end{figure*}

When domain shifts occur between the source and target domains, the performance of deep learning-based methods often degrades when tested on the target domain. If we confront domain shifts and data shortage simultaneously, it is like adding insult to injury. Suppose that we are given access to the hazy images in the target domain; we compare the results of the aforementioned methods on domain adaptation in Fig. \ref{fig: DE_DS}(a). We train the network on the partitions of Dense-Haze \cite{Dense-Haze_2019} and evaluate it on NH-Haze \cite{NH-Haze_2020} to simulate the scenario we have described. DAMix achieves the best results with respect to all fractions of the Dense-Haze dataset. Specifically, DAMix is again the most progressive data augmentation in the low data regime ($10\%$). Moreover, the network trained with DAMix using only $50\%$ of Dense-Haze performs even better than that trained with the whole Dense-Haze. We attribute these results to the design of DAMix and compliance with the atmospheric scattering model (\ref{Eq: ASM}), which allows DAMix to generate high-quality samples to mitigate the domain discrepancy between the source and target domains.

\subsubsection{Domain Generalization.}
We have demonstrated that DAMix dominates over all settings on domain adaptation. In this section, we study the performance on domain generalization, where hazy images in the target domain are not available. Since we are now unable to access the target domain, DAMix and CycleGAN \cite{CycleGAN2017} are typically not applicable. However, by randomizing $\mu^{t}$ (more details are provided in the supplementary material), DAMix can be applied to domain generalization. The experiments are performed by training the network on O-Haze \cite{O-HAZE_2018} and Dense-Haze \cite{Dense-Haze_2019} and evaluating it on NH-Haze \cite{NH-Haze_2020}. As shown in Fig. \ref{fig: DE_DS}(b), Copy-Blend \cite{shyam2021evaluating} fails to improve the performance in some settings ($10\%, 100\%$). In contrast, DAMix achieves universal improvements over all fractions of the training data. These results indicate that DAMix can not only be applied to domain adaptation, by randomizing $\mu^{t}$, DAMix can also improve the generalizability of the network. Moreover, DAMix is easy-to-use and does not have substantial computational overhead. DAMix can be easily plugged into any codebase to achieve better generalization.

\subsection{Adaptation to Real Images}
\label{sec: Real}
\begin{figure}[ht]
    \centering
    \includegraphics[width=.95\linewidth]{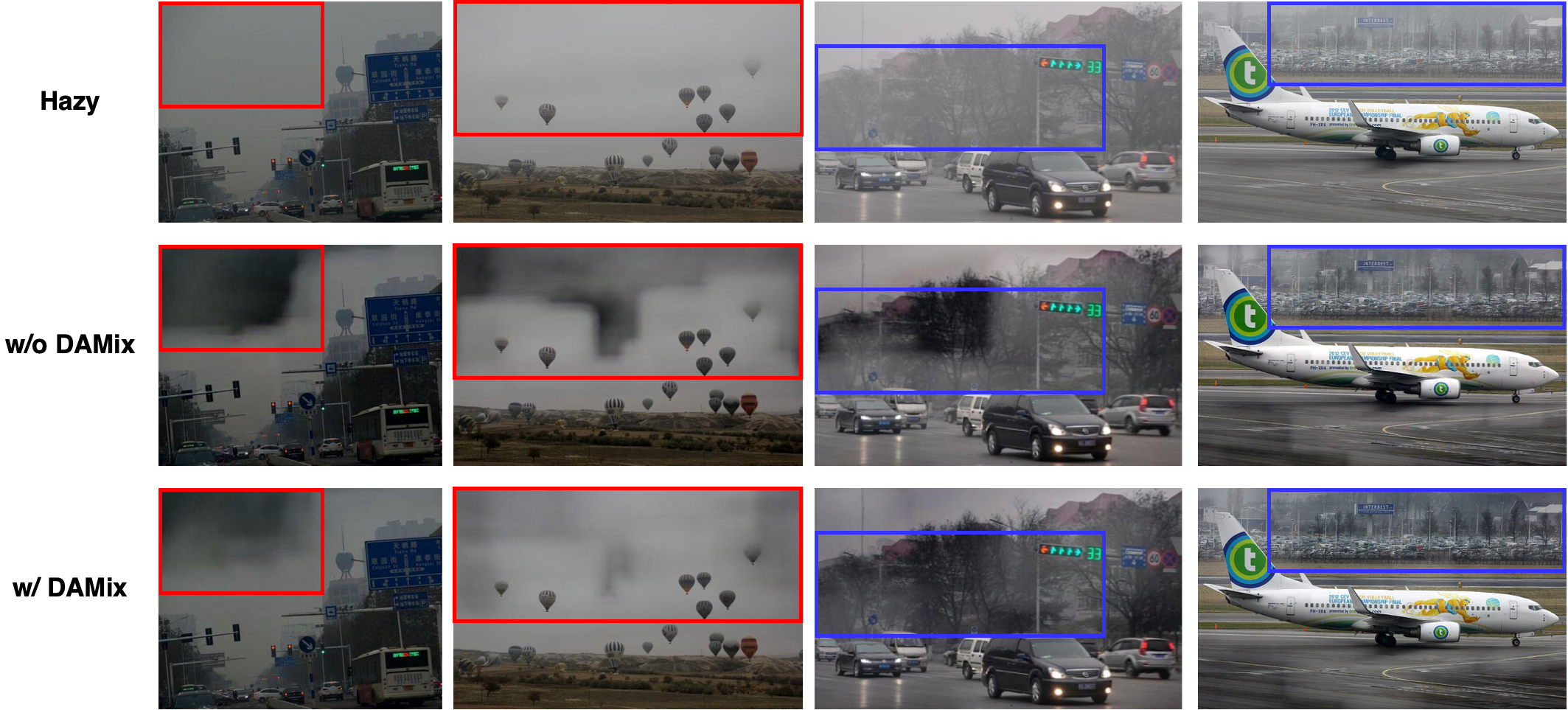}
    \caption{Effectiveness on the real-world dataset. The regions bounded by a red box show the ability of DAMix to mitigate color shifts in certain regions. The regions bounded by a blue box demonstrate that DAMix improves the ability of the network to handle thicker haze.}
    \label{fig: URHI}
\end{figure}

We have presented several results in Sec. \ref{sec: DA} and \ref{sec: DE} demonstrating that DAMix is superior to previously reported methods \cite{CycleGAN2017,shyam2021evaluating} when confronting domain shifts, regardless of effectiveness or efficiency. However, we have not yet applied DAMix to a real-world dataset. In this section, we train the network on OTS and evaluate it on URHI. As shown in Fig. \ref{fig: URHI}, the network trained without DAMix yields a color shift in the sky. In contrast, DAMix alleviates this problem to a certain extent. 
Furthermore, DAMix restores more details and improves the contrast in regions covered by thicker haze. We attribute these results to the diversity of the DAMix-ed samples, which endow the synthetic dataset with more variety in terms of haze distribution. Consequently, the network trained with DAMix provides more robust results when tested on the real-world dataset.

\subsection{Ablation Study}
Here, we verify the effectiveness of the DAMix formulation in Sec. \ref{sec: DAMix_form} by comparing the augmentation performance based on two settings. Setting \circled{1} employs the scalar combination in Eq. (\ref{Eq: EasyDAMix}) and applies the scalar-valued function (arithmetic mean of $\mathbf{I}_b$) to estimate the haze density. In contrast, setting \circled{2} is based on the formulation in Eq. (\ref{Eq: DAMix}), and the haze density is estimated according to Eq. (\ref{Eq: mu_I}). We conduct experiments under $D \rightarrow O$ and $D \rightarrow N$ to examine the difference between the two settings. As shown in Tab. \ref{tab: ablation_study}, setting \circled{1} and \circled{2} both surpass the baseline with flying colors in $D \rightarrow O$. However, for $D \rightarrow N$, only setting \circled{2} achieves considerable improvement. We attribute these results to three reasons. First, NH-Haze features a nonhomogeneous haze distribution, whereas O-Haze and Dense-Haze preserve high homogeneity. Second, because setting \circled{1} is subject to the scalar combination, it cannot modify the homogeneity of the source hazy images in Dense-Haze. In contrast, Eq. (\ref{Eq: DAMix}) endows setting \circled{2} with more degrees of freedom to better simulate the homogeneity in NH-Haze. Third, Eq. (\ref{Eq: mu_I}) can better interpret the haze distribution in NH-Haze. By contrast, the arithmetic mean of $\mathbf{I}_b$ can only describe the haze density but not the homogeneity of the haze. Accordingly, we suggest that setting \circled{2} is superior to setting \circled{1} in all respects when it comes to the formulation of DAMix.
\begin{table}[!ht]
    \centering
    \begin{tabular}{ccc}
        \toprule
        Setting & $D \rightarrow O$ & $D \rightarrow N$\\
        \midrule
        w/o Aug. & 16.11/0.579 & 15.78/0.593\\
        \circled{1} & 19.84/0.685 & 15.81/0.583\\
        \circled{2} & \textbf{20.22/0.686} & \textbf{16.53/0.608}\\
        \bottomrule
    \end{tabular}
    \caption{Ablation study on DAMix.}
    \label{tab: ablation_study}
\end{table}

\section{Conclusion}
Domain shifts have imposed a burden on several state-of-the-art (SOTA) dehazing methods. In this paper, we introduce the DAMix data augmentation method for mitigating the discrepancy between domains. DAMix has low computational overhead compared to previous work \cite{CycleGAN2017}, yet it is surprisingly effective for various deep learning-based methods. We have conducted thorough experiments on several SOTA methods \cite{liu2019griddehazenet,qin2020ffa,dong2020multi} and settings to verify the effectiveness and robustness of DAMix. DAMix dramatically improves the domain adaptation performance of these methods both quantitatively and qualitatively. Moreover, we have studied the potency of DAMix with respect to data efficiency under three different perspectives. The experimental results show that DAMix is superior to prior works \cite{CycleGAN2017,shyam2021evaluating} in general. Specifically, in terms of data efficiency for domain adaptation, DAMix performs best. We also demonstrate that DAMix can be applied to synthetic datasets to achieve better generalization on real-world datasets. Finally, we perform an ablation study to examine the rationality of the DAMix formulation and the haze density estimation. We hope that the convincing results will make DAMix a standard augmentation procedure for image dehazing when confronting domain shifts in practice.

{\small
\bibliographystyle{ieee_fullname}
\bibliography{ieee_fullname}
}

\end{document}